\documentclass{article}
\usepackage{graphicx}
\usepackage{wrapfig}
\usepackage{amsmath}
\usepackage{amssymb}
\usepackage{svg}

\usepackage{amsmath}
\usepackage{amssymb}
\usepackage{booktabs}
\usepackage{subcaption}
\usepackage{lipsum}
\usepackage{comment}
\usepackage{float}
\usepackage{wrapfig}
\usepackage[ruled,vlined]{algorithm2e}
\usepackage{xcolor}
\newcommand{\algblockblue}[1]{{\colorbox{blue!10}{\strut \textit{#1}}}}
\newcommand{\algblockred}[1]{{\colorbox{red!10}{\strut \textit{#1}}}}
\newcommand{\algblockgreen}[1]{{\colorbox{green!10}{\strut \textit{#1}}}}
\newcommand{\NAME}{\textsc{PairedGTA}}
\usepackage{subcaption}
\usepackage{soul}
\usepackage[a4paper,vmargin={20mm,20mm},hmargin={20mm,20mm}]{geometry}

\usepackage[utf8]{inputenc} 
\usepackage[T1]{fontenc}    
\usepackage{hyperref}       
\usepackage{url}            
\usepackage{booktabs}       
\usepackage{amsfonts}       
\usepackage{nicefrac}       
\usepackage{microtype}      
\usepackage{xcolor}         

\title{\textsc{PairedGTA}: Generating Driving Datasets for Controlled Photometric Shift Analysis}

\author{%
  Andrea Chianese$^{1}$, Giulio Rossolini$^{1}$, Alessandro Biondi$^{1}$,\\
  Marco Cococcioni$^{2}$, Giorgio Buttazzo$^{1}$\\[1.5em]
  $^{1}$Scuola Superiore Sant'Anna, Pisa, Italy, \\ Department of Excellence in Robotics \& AI, \texttt{\{name.surname\}@santannapisa.it}\\[0.5em]
  $^{2}$University of Pisa, DII, Pisa, Italy, 
  \texttt{marco.cococcioni@unipi.it}
}

\date{}

\begin{document}

\maketitle

\begin{abstract}
Evaluating the performance of visual perception systems for autonomous driving is essential to ensure reliable operation across diverse environmental scenarios. Ideally, a balanced and fair analysis across different adverse conditions would require perfectly paired images of the same scene under different weather or illumination changes. 
This would allow evaluating the effect of photometric shifts independently of geometry and semantic changes.
Unfortunately, real-world datasets rarely provide images of the same scene under different environmental conditions, because, normally, camera pose, traffic, and locations of dynamic objects (vehicles, pedestrians, etc.) vary over time, thus yielding only coarsely paired data.
To address this challenge, this work introduces a data generation framework based on a high-fidelity game engine for extracting perfectly paired images. By leveraging software APIs that communicate with the GTA game engine, the framework modifies illumination and weather conditions while preserving scene geometry, camera pose, and the identity and placement of dynamic objects. For each sampled location, it procedurally instantiates dynamic entities and renders pixel-aligned images under diverse adverse conditions.
The benefit of the proposed generation framework in driving scenarios is demonstrated through a systematic analysis of semantic segmentation models, whose output degradation can be attributed more directly to photometric shifts rather than to uncontrolled semantic or geometric factors. 
\end{abstract}


\section{Introduction}

\begin{figure}[t]
        \centering
        \includegraphics[width=0.7\linewidth]{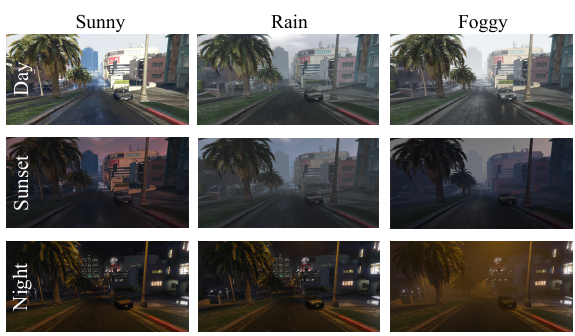}
        \caption{Examples of paired images produced by the proposed framework. Dynamic objects are generated once and consistently preserved across multiple photometric conditions, enabling pixel-aligned comparisons under controlled environmental changes.}
        \label{fig:intro_figure}
\end{figure}

In autonomous driving, reliable deployment of visual perception systems, such as semantic segmentation models, requires not only high average accuracy, but also a precise understanding of how and why performance degrades under adverse environmental conditions, such as nighttime, rain, or fog~\cite{zendel2018kitti,neuhold2017mapillary}. Among driving benchmarks, particular attention has been paid to paired datasets, where the same scenes are observed under different conditions~\cite{sakaridis2021acdc}. In this setting, perfect pairing across conditions is essential for a reliable self-supervised analysis,
where ground truth is often scarce.
In fact, without ground truth, two serious issues emerge. First, imperfect pairing limits the possibility of leveraging output correspondences across conditions, which are crucial for learning and testing without pixel-level labels.
Second, the observed performance gap may reflect not only changes in photometric conditions, such as illumination or weather, but also geometric and semantic differences in scene geometry, traffic composition, object presence, and occlusions. 

In practice, real-world datasets rarely provide exact correspondence across conditions, since scenes inevitably change over time~\cite{torralba2011dataset,ben2010theory}.
Generative approaches attempt to address this limitation through image-to-image translation across weather or illumination conditions~\cite{anoosheh2018todaygan,gella2023weatherproof}, but they may introduce artifacts, unintended texture changes, or fail to guarantee pixel-level correspondence between pairs. Open simulators allow a much better control over rendering conditions~\cite{ha2016synthia,doso2017carla,sun2022shift, syndra2025damico}; however, building complex photorealistic environments often requires substantial effort and costly design to achieve realism.

Inspired by the extensive use of game engines in driving vision tasks~\cite{richter2016playing,richter2017playing,presil2016,kiefer2021leveragingsyntheticdataobject,martinez2017grandtheftautov,cao2022datagenerationusingsimulation}, this work introduces \NAME, a dataset generation framework based on GTA~V~\cite{gta}
for generating paired data. As illustrated in Figure \ref{fig:intro_figure}, the framework intervenes on illumination and weather while preserving scene geometry, camera pose, and the identity and placement of dynamic objects across all generated variants. Starting from a canonical scene configuration, it procedurally instantiates dynamic entities, including vehicles, pedestrians, and other traffic participants, and renders multiple photometric variants from the same viewpoint. This enables pixel-aligned comparisons in which non-photometric factors remain fixed, allowing model behavior to be analyzed under controlled environmental changes.

The benefit and efficiency of the framework are demonstrated through a set of experiments aimed at testing the reliability of different models across diverse conditions using self-consistency metrics. We further show that such paired data provide a stronger understanding of cross-condition consistency than real-world adverse-condition datasets, where weather and illumination changes are entangled with variations in traffic composition, object positions, and scene content. We believe this framework can support a large-scale generation of paired data and
advance the use of self-supervised analysis for both training and testing models for visual perception in autonomous driving. The code of the framework, experimental analysis, and dataset are publicly available at [\footnote{\url{https://github.com/Spearton/PairedGTA}}]. 

\paragraph{Paper contributions.}
In summary, this work provides the following contributions:
\begin{itemize}
    \item It introduces a deterministic game-engine-based framework for generating pixel-aligned driving scenes under controlled illumination and weather conditions.
    \item It proposes a counterfactual photometric robustness analysis that preserves geometry, camera pose, and dynamic object identity across all generated variants.
    \item It presents a set of experimental results showing that controlled photometric datasets provide clearer insights for an output-comparative testing of segmentation models than real-world adverse-condition datasets, where photometric, semantic, and geometric factors are entangled.
\end{itemize}

\paragraph{Paper organization.}
The remainder of the paper is organized as follows. Section~\ref{sec:rel} reviews the related literature and discusses the motivation behind this work. Section~\ref{sec:framework} introduces the proposed framework, describing the generation pipeline and its constraints. Section~\ref{sec:experiments} reports the experimental results on semantic segmentation models. Finally, Section \ref{sec:conclusion} states the conclusions. 

\section{Related work}
\label{sec:rel}
This section reviews prior work on driving datasets and data generation frameworks, then discusses the role of paired images for controlled robustness analysis, and finally positions our contribution with respect to real-world, simulation-based, and generative approaches.

\paragraph{Driving datasets and generation frameworks.}
Driving datasets are crucial for the development and evaluation of visual perception models in autonomous driving. Real-world benchmarks such as Cityscapes~\cite{cordts2016cityscapes}, Mapillary Vistas~\cite{neuhold2017mapillary}, and BDD100K~\cite{yu2020bdd100k} provide diverse urban scenes and dense annotations for tasks such as semantic segmentation, object detection, and scene understanding. However, models trained and evaluated on standard driving datasets may still suffer significant degradation under adverse environmental conditions, such as nighttime, rain, fog, or snow. This has motivated the introduction of adverse-condition and corruption-oriented benchmarks~\cite{sakaridis2021acdc,sakaridis2019guided, sakaridis2018semantic}, and broader robustness benchmarks based on synthetic corruptions and natural distribution shifts~\cite{hendrycks2019benchmarking,michaelis2019benchmarking,taori2020measuring}.

In parallel, synthetic and simulation-based frameworks have been widely adopted to generate annotated driving data with controllable conditions. Datasets and simulators such as SYNTHIA~\cite{ha2016synthia}, Virtual KITTI~\cite{gaidon2016virtual}, CARLA~\cite{doso2017carla}, and SHIFT~\cite{sun2022shift} enable the creation of driving scenes under different weather, illumination, and domain configurations. These platforms provide strong control over scene content and annotations, but generating large-scale data in complex and realistic environments may require substantial effort, especially when detailed urban layouts, dynamic actors, and diverse traffic situations must be manually designed or scripted.
To alleviate part of this effort while preserving a high level of visual realism, commercial game engines have been extensively exploited to obtain large-scale photorealistic driving data. Some GTA-based works~\cite{richter2016playing, richter2017playing} demonstrated the potential of game environments for generating realistic annotated data at scale. These synthetic datasets have been valuable for both training and evaluating perception models, including domain adaptation from synthetic to real driving scenes~\cite{hoffman2016fcns, tsai2018learning, sankaranarayanan2018learning}. Nevertheless, most existing datasets and generation frameworks are primarily designed to increase data diversity, coverage, or realism, rather than enforcing balanced adverse-condition variants or strict correspondence between the same scene rendered under multiple environmental conditions.

\paragraph{Paired images for controlled robustness analysis.}
The above limitation is particularly relevant when the goal is to evaluate the robustness of a model in tolerating adverse conditions due to illumination and weather changes. In this case, paired images are essential for a precise robustness analysis, since they allow comparing the model behavior across different environmental conditions while preserving the underlying scene semantics. In an ideal paired setting, the same objects, geometry, camera pose, and spatial layout are observed under different photometric conditions. This allows understanding whether prediction changes are caused by the environmental shift itself or by a different scene content.
 The consistency of the two outputs produced on the paired images can serve as an unsupervised measure of the model's robustness, especially when dense annotations are unavailable for every condition.

However, obtaining perfectly paired images in real-world scenarios is challenging. Even when the same location is revisited, variations in camera pose, traffic composition, object placement, occlusions, and pedestrian or vehicle presence can alter the scene. As a result, the observed performance gap may not be caused only by photometric changes, but also by semantic and geometric differences. Real-world paired datasets such as ACDC~\cite{sakaridis2021acdc} and Dark Zurich~\cite{sakaridis2019guided} provide a significant step toward cross-condition analysis, but they cannot fully guarantee counterfactual correspondence across conditions for the above reasons.

To address these challenges, generative approaches have been explored to obtain paired or condition-translated images. Image-to-image translation and recent diffusion-based methods can synthesize adverse weather or illumination changes from a given input image~\cite{anoosheh2018todaygan, gella2023weatherproof, weatherflux2025, pix2pix_weather2025, jia2024dginstyle}. These approaches are useful for data augmentation, domain adaptation, and domain generalization. However, generative models may still behave unpredictably and introduce artifacts, modify textures, alter object boundaries, or change local structures, as commonly observed in image-to-image translation pipelines~\cite{isola2017image,zhu2017unpaired}. These effects limit their suitability for strict pixel-level comparative analysis, where even small structural changes may confound the interpretation of model behavior.

\paragraph{Positioning of this work.}
This work addresses the need for controlled paired data specifically designed for photometric shift analysis. It builds on game engines for generating deterministic multi-condition variants of the same scene.
Unlike generative translation methods, it does not synthesize adverse conditions through stochastic image manipulation, but renders each condition directly from the same scene configuration. This provides a deterministic form of paired generation, where camera pose, scene geometry, and identity and placement of dynamic objects are preserved across conditions.

\section{Methodology}
\label{sec:framework}
This section introduces the proposed data generation methodology. The main objective is to generate groups of images representing the same underlying driving scene under different environmental conditions, while preserving camera pose, scene geometry, and dynamic object configuration. From an implementation perspective, the framework relies on a client-server interface to GTA V that enables scene initialization, environmental control, dynamic object manipulation, and repeated rendering of the same scene under different photometric conditions.

\subsection{Definition of a photometric-shifted dataset}

We formalize a photometric-shifted dataset as a collection of image groups, where each group contains multiple renderings of the same scene under different illumination or weather conditions. Let
\(
\mathcal{C}_{\text{photo}} = \{c_i\}_{i=1}^{N}
\)
be the set of photometric conditions considered in the dataset. Each condition \(c_i\) specifies environmental rendering parameters, such as time of day, illumination, rain, fog, or cloud coverage.
For each scene \(k\), the framework first constructs an internal scene descriptor \(\mathcal{S}_k\). This descriptor specifies all non-photometric properties that must remain fixed across conditions, including the camera pose, the static scene layout, and the configuration of dynamic objects. The descriptor is not part of the final dataset, but it is used by the framework to render the aligned image variants.

Let \(R\) denote the rendering function that generates an image from a scene descriptor and a photometric condition. The image associated with scene \(k\) and condition \(c_i\) is therefore given by
\(
    \mathbf{X}_{k,i} = R(\mathcal{S}_k, c_i).
\)
The group of images generated from the same scene descriptor is defined as
\(
    \mathcal{X}_k = \left\{ \mathbf{X}_{k,i} \right\}_{i=1}^{N}.
\)
The complete dataset is then defined as
\(
    \mathcal{D} = \left\{ \mathcal{X}_k \right\}_{k=1}^{K}.
\)
Thus, each dataset element \(\mathcal{X}_k\) contains multiple photometric variants of the same scene. Since all images in \(\mathcal{X}_k\) are generated from the same descriptor \(\mathcal{S}_k\), differences among them are induced only by changes in the photometric conditions.

Note that, in our setting, the game-based environment is used to generate controlled visual variants of the same scene, but it does not provide direct access to ground-truth labels. This is a practical limitation of several game-based data generation pipelines \cite{richter2016playing, presil2016, kiefer2021leveragingsyntheticdataobject}, where the simulator can be controlled and queried through external APIs, while the internal semantic labels of the rendered scene are not directly exposed. A more detailed discussion of this aspect is provided in Section~\ref{sec:conclusion}. For this reason, the experimental analysis in Section~\ref{sec:experiments} does not rely on absolute performance measured against simulator-provided ground truth. Instead, we use pseudo-labels obtained from high-confidence predictions of the most reliable model in the non-adverse scenario, corresponding to sunny daytime, as the reference.

\subsection{Data generation}
The data generation process consists of three stages, which are summarized in Algorithm~\ref{alg:dataset_generation}: \emph{(i)} location sampling and scene initialization; \emph{(ii)} semantic scene construction; and \emph{(iii)} condition-wise rendering. These stages are described below and define how each image group \(\mathcal{X}_k\) is generated.

\begin{wrapfigure}{r}{0.48\textwidth}
\vspace{-2em}
\begin{algorithm}[H]
\caption{Data Generation}
\label{alg:dataset_generation}
\DontPrintSemicolon
\footnotesize
\BlankLine
$\mathcal{D} \gets \emptyset$\;
Initialize $\mathcal{C}_{\text{photo}}$\;
\For{$k \gets 1$ \KwTo $K$}{
    \algblockblue{Location sampling and scene initialization} \\
    Sample a valid location $\mathbf{p}_k$\;
    Initialize vehicle and camera pose at $\mathbf{p}_k$\;
    \BlankLine
    \algblockred{Semantic scene construction} \\
    Instantiate dynamic objects $\mathcal{O}_k$ around $\mathbf{p}_k$\;
    Build scene descriptor $\mathcal{S}_k$\;
    \BlankLine
    \algblockgreen{Condition-wise rendering} \\
    $\mathcal{X}_k \gets \emptyset$\;
    \ForEach{$c_i \in \mathcal{C}_{\text{photo}}$}{
        Initialize scene from $\mathcal{S}_k$\;
        Apply photometric condition $c_i$\;
        $\mathbf{X}_{k,i} \gets \text{CaptureScene}(\mathcal{S}_k, c_i)$\;
        $\mathcal{X}_k \gets \mathcal{X}_k \cup \{\mathbf{X}_{k,i}\}$\;
    }
    \BlankLine
    $\mathcal{D} \gets \mathcal{D} \cup \{\mathcal{X}_k\}$\;
}
\Return{$\mathcal{D}$}
\end{algorithm}
\end{wrapfigure}

\paragraph{Location sampling and scene initialization.}
A location identifies the spatial position in the virtual world around which a driving scene is generated. For each scene \(k\), the framework samples a location
\(
\mathbf{p}_k = (x_k, y_k),
\)
corresponding to the ego-vehicle's position on the virtual world's ground plane. Locations are sampled from a predefined valid region covering the metropolitan area and some selected peripheral zones, while excluding invalid areas such as bodies of water or non-navigable regions.

After selecting a location \(\mathbf{p}_k\), the simulator is initialized at the corresponding position. In this context, initialization means setting the ego vehicle and camera pose, clearing the local simulation state, and preparing the environment before dynamic objects are instantiated. The first photometric condition \(c_1 \in \mathcal{C}_{\text{photo}}\) (which is sunny day) is used as the default rendering condition for scene construction.

\begin{figure}[t]
    \centering
    \includegraphics[width=\linewidth]{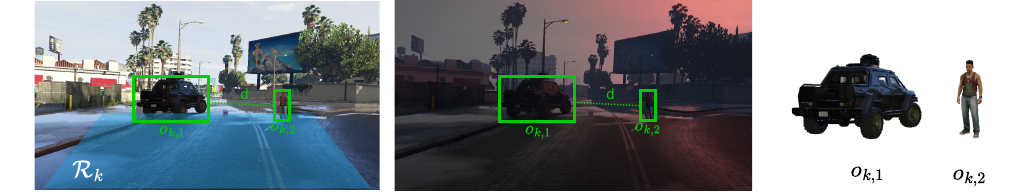}
    \caption{\small{Example of a daytime sunny image and a corresponding sunset sunny variant generated by the framework, highlighting object poses and placement constraints.
    }}
    \label{fig:placement_illustration}
\end{figure}

\paragraph{Semantic scene construction.}
After initialization, the framework constructs the semantic scene associated with scene \(k\). A set of dynamic objects
\(
\mathcal{O}_k = \{o_{k,1}, \dots, o_{k,M_k}\}
\)
is instantiated around the sampled location \(\mathbf{p}_k\). 
Each object \(o_{k,j}\) is drawn from a predefined category set, such as car, motorcycle, or pedestrian, and assigned a position \(\mathbf{q}_{k,j}\). To avoid physically implausible overlaps, objects cannot be placed at a distance less than $d_{min}$, that is, their positions satisfy the constraint
\(
\|\mathbf{q}_{k,j} - \mathbf{q}_{k,l}\|_2 \geq d_{\min}, \forall j \neq l.
\)
Object positions are not sampled over the entire virtual world. Instead, they are drawn from a bounded region \(\mathcal{R}_k\) centered around the ego vehicle and constrained to valid road or sidewalk regions, depending on the object category. This favors plausible object placement with respect to the ego-vehicle orientation and camera field of view, while maintaining variability across generated scenes. An illustration of this placement process is shown in Figure~\ref{fig:placement_illustration}.

Once the dynamic objects have been instantiated, the full non-photometric configuration is stored in the scene descriptor \(\mathcal{S}_k\). This descriptor includes the camera pose, the static scene layout, and the complete dynamic object configuration, including object categories, identities, and positions. 

\paragraph{Condition-wise rendering.}
For each of the photometric condition $c_i \in \mathcal{C}_{\text{photo}}$, the simulator uses the same scene descriptor $\mathcal{S}_k$, re-initializes the scene at the same location $\mathbf{p}_k$
and camera pose, applies the selected condition $c_i$, adds dynamic objects, and captures the corresponding image $\mathbf{X}_{k,i}$. Repeating this process for all conditions in $\mathcal{C}_{\text{photo}}$ yields the aligned image group $\mathcal{X}_k = \{\mathbf{X}_{k,i}\}_{i=1}^{N}$.
By construction, all images in \(\mathcal{X}_k\) share the same geometry, camera pose, and dynamic object layout, while differing only in illumination and weather. 

\subsection{Software details and APIs}
\label{sec:techcnical_details}

\begin{figure*}[htbp]
    \centering
    \includegraphics[width=\linewidth]{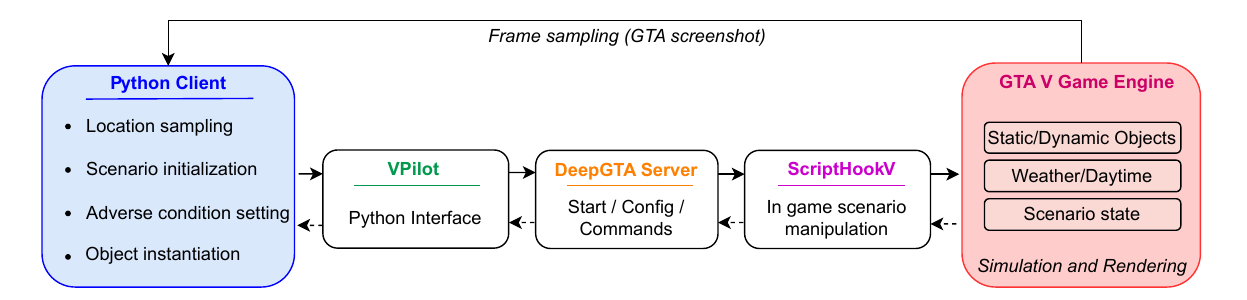}
    \caption{\small{Communication pipeline between the proposed framework (blue block), other third-party software tools, and the GTA game engine.
    }}
    \label{fig:met_schema}
\end{figure*}

The proposed data generation pipeline is built on top of the \textit{DeepGTAV} framework~\cite{frameworkDeepGTA}, which provides a communication layer between the GTA V game engine and an external client. As illustrated in Figure~\ref{fig:met_schema}, the generation process is orchestrated by a Python client, which is responsible for location sampling, scenario initialization, adverse-condition configuration, and dynamic object instantiation.

The interaction with the game engine is mediated by \textit{VPilot}~\cite{frameworkVpilot}, which provides a Python interface for client-side communication and scenario orchestration. VPilot communicates with the DeepGTAV server, exposed by the game plugin through a TCP connection on port 8000, by default. Through this interface, the client can send \textit{Start} and \textit{Config} messages to initialize the environment and define the desired acquisition settings, \textit{Commands} messages to control the ego vehicle when required, and \textit{Stop} messages to terminate the simulation session and return to normal gameplay.

Once the environment has been initialized, DeepGTAV streams the requested data back to the client in JSON format. Depending on the selected configuration, this information may include rendered frames, environmental parameters, surrounding vehicles and pedestrians, and additional metadata required for dataset generation. The framework uses these data streams to acquire condition-specific views of the same underlying scene, while preserving the desired camera pose, geometry, and semantic layout.

At a lower level, the pipeline relies on \textit{Script Hook V}~\cite{scriptScriptHookV}, which exposes GTA V native scripting functions to custom \texttt{.asi} plugins. This component enables fine-grained manipulation of the game state, including static and dynamic objects, weather and time settings, and scenario-related parameters. These low-level controls allow the framework to modify the simulated environment beyond the default DeepGTAV configuration options, supporting the controlled generation of paired photometric variants.

\section{Experiments}
\label{sec:experiments}

This section evaluates the proposed generation framework using semantic segmentation by analyzing the consistency of model predictions under controlled photometric shifts. 

\subsection{Experimental setup}
\begin{wrapfigure}{r}{0.50\textwidth}
        \vspace{-1em}
        \centering
        \includegraphics[width=0.50\textwidth]{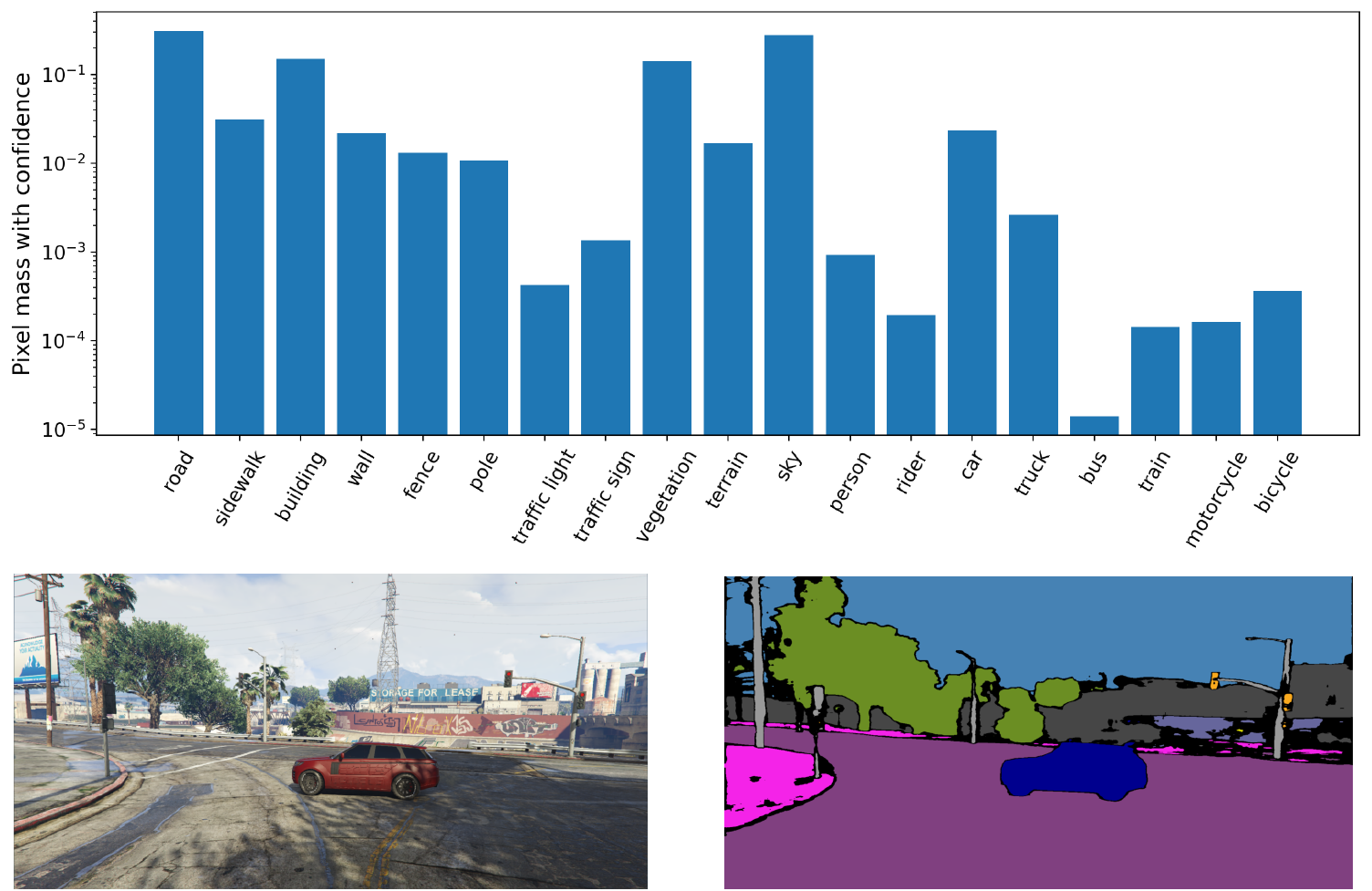}
    \caption{\small{Dataset-level distribution of pseudo-labels generated with SegFormer-B5. The bottom panels show an example of a reference image and the segmentation map.}}
    \label{fig:distribution_pseudo_labels}
    \vspace{-1em}
\end{wrapfigure}
\paragraph{Dataset.}
The evaluation is conducted on a dataset generated using the pipeline described in Section~\ref{sec:framework}. The dataset contains more than 100 unique spatial locations. For each location $k$, we generate nine spatially aligned images corresponding to all combinations of three illumination conditions, namely \textit{Day}, \textit{Sunset}, and \textit{Night}, and three weather conditions, namely \textit{Sunny}, \textit{Rain}, and \textit{Foggy}. For each location, we additionally insert a random number of dynamic objects between 0 and 3.
As mentioned in Section~\ref{sec:framework}, the game-based environment does not provide direct access to dense semantic ground-truth annotations. 
Therefore, for each image group $\mathcal{X}_k = \{\mathbf{X}_{k,i}\}_{i=1}^{N}$, pseudo-labels are derived by segmenting the \textit{Day--Sunny} image $\mathbf{X}_{k,0}$ by the SegFormer-B5 model~\cite{segformer}.
Only pixels predicted with a confidence higher than a threshold $\tau=0.6$ are retained as valid pseudo-labels. Pixels with confidence below this threshold are assigned to a void class and are excluded from the evaluation metrics. This produces a high-confidence pseudo-label map $\tilde{\mathbf{y}}_{k}$. 
Figure~\ref{fig:distribution_pseudo_labels} reports the distribution of pixel mass across the generated dataset and shows an example of the pseudo-labels extracted from a reference scenario.

\paragraph{Models.}
The experiments consider a diverse set of semantic segmentation architectures trained on Cityscapes~\cite{cordts2016cityscapes}, covering both transformer-based and convolutional designs. In particular, the evaluation includes SegFormer variants B0, B2, and B5~\cite{segformer}, Mask2Former with Swin-S and Swin-L backbones~\cite{mask2former}, DeepLabV3+ with a ResNet-101 backbone~\cite{chen2018deeplabv3plus}, PIDNet-M~\cite{xu2023pidnet}, and DDRNet-23~\cite{hong2021ddrnet}. These models provide a representative comparison across high-capacity transformer-based methods, encoder--decoder convolutional networks, and efficient real-time segmentation architectures.

\paragraph{Evaluation protocol and metrics.}
Since the simulator does not provide dense ground truth, the consistency of model predictions is evaluated with respect to the pseudo-labels extracted by semantic segmentation.
For a given image group $k$, let $\tilde{y}_{k}^{(u)}$ denote the pseudo-label assigned to pixel location $u$ in the clean reference image. Pixels whose confidence is below the threshold $\tau$ are mapped to the void class $\varnothing$ and are excluded from the evaluation.
Given a segmentation model $f$, let
\(
    \hat{y}_{k,i}^{(u)} = f(\mathbf{X}_{k,i})^{(u)}
\)
be the predicted class at pixel location $u$ for the image generated at location $k$ under condition $i$, where $i=0$ corresponds to the clean \textit{Day--Sunny} reference condition. For each semantic class $c \neq \varnothing$, the set of valid reference pixels is defined as
\(
    \Omega_{k,c} =
    \left\{
    u :
    \tilde{y}_{k}^{(u)} = c
    \right\}.
    \label{eq:valid_reference_pixels}
\)
Then, the class-wise reference consistency score is computed as
\begin{equation}
    \xi_{c}(i) =
    \frac{
    \sum_{k} \sum_{u \in \Omega_{k,c}}
    \mathbf{1}\!\left[
    \hat{y}_{k,i}^{(u)} = c
    \right]
    }{
    \sum_{k} |\Omega_{k,c}|
    }.
    \label{eq:class_reference_consistency}
\end{equation}
where $\mathbf{1}[\cdot]$ denotes the indicator function, which is equal to $1$ when the condition inside brackets is true and $0$ otherwise.
When $\xi_{c}(i)$ is close to $1$ indicates that pixels assigned to class $c$ in the clean reference condition are still predicted as class $c$ under the shifted condition. Conversely, lower values indicate that the photometric shift induces prediction changes on pixels that were confidently assigned to class $c$ in the reference image. Since the metric is computed with respect to pseudo-labels extracted from the clean reference image, it should be interpreted as a reference-based consistency score rather than as absolute segmentation accuracy.

\subsection{Evaluating photometric shifts on segmentation models}
\begin{table*}[t]
\centering
\renewcommand{\arraystretch}{0.9}
\begin{subtable}[t]{0.95\textwidth}
\centering
\caption{Day-Sunny $\rightarrow$ Day-Rain.}
\label{tab:retention_day_rain}
\scriptsize
\setlength{\tabcolsep}{3pt}
\resizebox{\linewidth}{!}{%
\begin{tabular}{lccccccccccccccccc|c}
\toprule
Model & road & sidewalk & building & wall & fence & pole & T. light & T. sign & vegetation & terrain & sky & person & rider & car & truck & motor. & bicycle & Mean \\
\midrule
DDRNet-23 & 56.3 & 13.8 & \textbf{93.8} & 0.5 & 0.6 & 27.0 & 12.1 & 9.9 & 84.8 & 0.2 & 66.2 & 62.3 & 75.2 & 74.4 & 0.0 & 53.0 & 0.0 & 37.1 \\
DeepLabV3+ R101 & 97.2 & 50.7 & 82.8 & 38.3 & 58.0 & 52.4 & 5.4 & 47.6 & 90.0 & 42.0 & 96.0 & 85.6 & 68.0 & 87.8 & 0.0 & 75.7 & 0.0 & 57.5 \\
Mask2Former Swin-L & 96.8 & 51.6 & 82.4 & 44.6 & 75.7 & 59.5 & 12.2 & 71.9 & 88.8 & \textbf{84.0} & 98.2 & 87.3 & 75.1 & 85.6 & 26.8 & 88.5 & 0.0 & 66.4 \\
Mask2Former Swin-S & 95.5 & 40.0 & 83.3 & 24.0 & 66.5 & 61.7 & 18.5 & 28.7 & 88.1 & 76.2 & 97.7 & 83.2 & 81.1 & 87.6 & \textbf{75.0} & 58.3 & 0.0 & 62.7 \\
PIDNet-M & 93.7 & 18.4 & 74.5 & 50.8 & 41.2 & 40.9 & 10.0 & 53.7 & 86.9 & 9.4 & 83.7 & 83.5 & 63.6 & 62.3 & 4.4 & 65.1 & 0.0 & 49.5 \\
SegFormer-B0 & 97.3 & 43.8 & 81.7 & 62.3 & 69.2 & 52.8 & 8.3 & 70.1 & 90.3 & 40.1 & 97.8 & 57.7 & 78.2 & 83.8 & 14.9 & 57.5 & 0.0 & 59.2 \\
SegFormer-B2 & \textbf{97.4} & 49.8 & 85.3 & 69.6 & 70.4 & 66.3 & 13.4 & 72.4 & 88.2 & 68.2 & 96.8 & 72.1 & \textbf{88.2} & \textbf{92.6} & 14.1 & 22.7 & 0.0 & 62.8 \\
SegFormer-B5 & 97.0 & \textbf{62.2} & 90.6 & \textbf{85.4} & \textbf{88.7} & \textbf{79.7} & \textbf{25.3} & \textbf{88.0} & \textbf{93.9} & 45.3 & \textbf{98.8} & \textbf{88.2} & 84.0 & 91.9 & 52.7 & \textbf{89.9} & \textbf{96.0} & \textbf{79.9} \\
\bottomrule
\end{tabular}
}
\end{subtable}
\begin{subtable}[t]{0.95\textwidth}
\centering
\caption{Day-Sunny $\rightarrow$ Sunset-Sunny.}
\label{tab:retention_sunset_sunny}
\scriptsize
\setlength{\tabcolsep}{3pt}
\resizebox{\linewidth}{!}{%
\begin{tabular}{lccccccccccccccccc|c}
\toprule
Model & road & sidewalk & building & wall & fence & pole & T. light & T. sign & vegetation & terrain & sky & person & rider & car & truck & motor. & bicycle & Mean \\
\midrule
DDRNet-23 & 10.9 & 1.4 & \textbf{92.5} & 0.0 & 19.2 & 16.4 & 49.1 & 24.6 & 73.9 & 0.5 & 0.6 & 36.4 & 65.8 & 92.5 & 22.0 & \textbf{0.0} & 0.0 & 29.8 \\
DeepLabV3+ R101 & 97.9 & 55.6 & 75.5 & 25.6 & 32.6 & 71.5 & 53.3 & 56.1 & 93.6 & 4.5 & 96.6 & 71.7 & 36.0 & 85.8 & 38.2 & \textbf{0.0} & 18.7 & 53.7 \\
Mask2Former Swin-L & \textbf{98.0} & 61.7 & 80.3 & 67.0 & 52.8 & 69.6 & 45.3 & 43.2 & 90.9 & 40.7 & 97.1 & 60.1 & \textbf{95.2} & 93.9 & 92.9 & \textbf{0.0} & 68.3 & 68.1 \\
Mask2Former Swin-S & 97.0 & 42.0 & 83.4 & 41.8 & 46.6 & 63.6 & 67.8 & 44.7 & 86.9 & \textbf{64.7} & 96.3 & 47.9 & 0.0 & 92.6 & 96.0 & \textbf{0.0} & 71.3 & 61.3 \\
PIDNet-M & 83.1 & 14.9 & 72.8 & 39.5 & 25.9 & 56.3 & 58.4 & 18.0 & 94.0 & 3.1 & 35.3 & 69.1 & 36.0 & 77.5 & 83.8 & \textbf{0.0} & 0.0 & 45.2 \\
SegFormer-B0 & 96.2 & 52.7 & 78.6 & 61.6 & 34.3 & 50.7 & 32.8 & 5.0 & 87.5 & 6.0 & 82.1 & 10.8 & 0.9 & 95.0 & 0.0 & \textbf{0.0} & \textbf{79.9} & 45.5 \\
SegFormer-B2 & 97.0 & 53.2 & 78.5 & \textbf{72.6} & 41.1 & 67.1 & 55.7 & 44.8 & 91.7 & 9.5 & 93.9 & 41.9 & 93.3 & 96.0 & 0.0 & \textbf{0.0} & 77.3 & 59.6 \\
SegFormer-B5 & 97.5 & \textbf{87.4} & 90.1 & 60.5 & \textbf{58.5} & \textbf{84.9} & \textbf{73.6} & \textbf{72.9} & \textbf{96.3} & 31.3 & \textbf{98.5} & \textbf{77.3} & 93.5 & \textbf{97.8} & \textbf{99.6} & \textbf{0.0} & 78.9 & \textbf{76.4} \\
\bottomrule
\end{tabular}
}
\end{subtable}
\begin{subtable}[t]{0.95\textwidth}
\centering
\caption{Day-Sunny $\rightarrow$ Night-Sunny.}
\label{tab:retention_night_sunny}
\scriptsize
\setlength{\tabcolsep}{3pt}
\resizebox{\linewidth}{!}{%
\begin{tabular}{lccccccccccccccccc|c}
\toprule
Model & road & sidewalk & building & wall & fence & pole & T. light & T. sign & vegetation & terrain & sky & person & rider & car & truck & motor. & bicycle & Mean \\
\midrule
DDRNet-23 & 34.2 & 4.9 & 78.8 & 11.9 & 4.2 & 14.6 & 1.0 & 0.2 & 75.2 & 1.7 & 0.0 & 46.1 & \textbf{0.0} & 27.6 & 0.0 & 0.0 & \textbf{0.0} & 17.7 \\
DeepLabV3+ R101 & 96.7 & 39.4 & 83.9 & 34.1 & 37.9 & 50.2 & \textbf{59.8} & 10.6 & 85.6 & 16.8 & 62.6 & 86.7 & \textbf{0.0} & 72.3 & 0.0 & 0.0 & \textbf{0.0} & 43.3 \\
Mask2Former Swin-L & 96.6 & 53.4 & 88.4 & 67.5 & 11.0 & 56.3 & 45.4 & 5.0 & 82.2 & \textbf{65.1} & \textbf{90.5} & 83.9 & \textbf{0.0} & 28.2 & 0.3 & \textbf{99.2} & \textbf{0.0} & 51.4 \\
Mask2Former Swin-S & 93.9 & 65.8 & 85.8 & 58.3 & 63.0 & 50.4 & 34.0 & 6.1 & 68.2 & 51.8 & 70.1 & 76.6 & \textbf{0.0} & 52.9 & 2.8 & 0.0 & \textbf{0.0} & 45.9 \\
PIDNet-M & 78.4 & 40.5 & 73.8 & 15.4 & 10.4 & 32.0 & 34.7 & 2.1 & 72.8 & 2.4 & 6.4 & 78.8 & \textbf{0.0} & 17.1 & 6.5 & 0.0 & \textbf{0.0} & 27.7 \\
SegFormer-B0 & 81.5 & 38.2 & 85.8 & 47.2 & 27.7 & 36.1 & 27.4 & 1.4 & 83.4 & 22.4 & 29.3 & 51.5 & \textbf{0.0} & 56.0 & 0.0 & 0.0 & \textbf{0.0} & 34.6 \\
SegFormer-B2 & \textbf{97.1} & 52.3 & 86.7 & 38.4 & 59.4 & 50.6 & 5.1 & 0.0 & 79.9 & 29.2 & 39.7 & 72.7 & \textbf{0.0} & 51.9 & 0.0 & 0.0 & \textbf{0.0} & 39.0 \\
SegFormer-B5 & 95.7 & \textbf{67.8} & \textbf{94.4} & \textbf{87.4} & \textbf{78.5} & \textbf{64.6} & 35.1 & \textbf{13.5} & \textbf{87.6} & 47.3 & 41.4 & \textbf{89.0} & \textbf{0.0} & \textbf{84.7} & \textbf{17.7} & 0.0 & \textbf{0.0} & \textbf{53.2} \\
\bottomrule
\end{tabular}
}
\end{subtable}
\vspace{1em}
\caption{\small{class-wise reference consistency across environmental conditions and multiple models. The last column reports the class-wise mean, and the best values are shown in bold.}}
\label{tab:retention_conditions}
\end{table*}

\begin{figure}[t]
\centering
    \begin{subfigure}{0.3\textwidth}
        \centering
    \includegraphics[width=\textwidth]{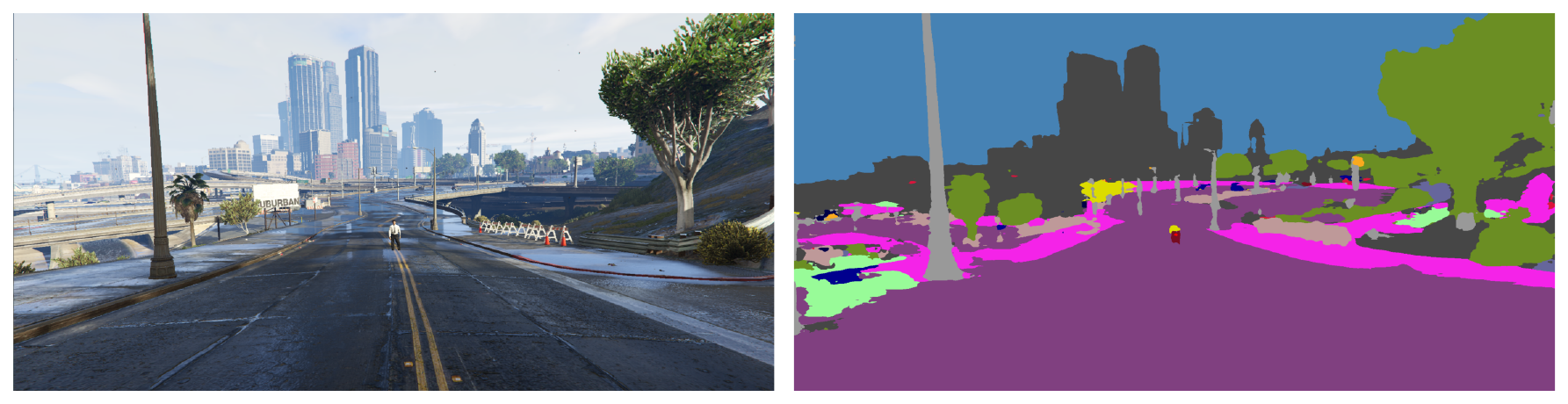}
    \end{subfigure}
    \begin{subfigure}{0.3\textwidth}
        \centering
        \caption*{\footnotesize SegFormer-B5}
    \includegraphics[width=\textwidth]{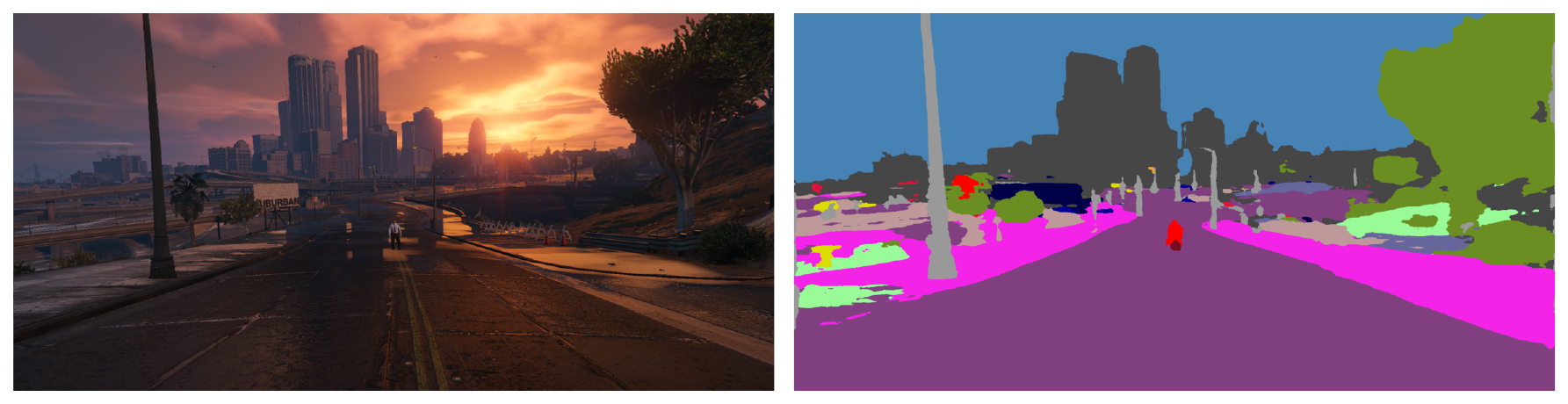}
    \end{subfigure}
    \begin{subfigure}{0.3\textwidth}
        \centering
    \includegraphics[width=\textwidth]{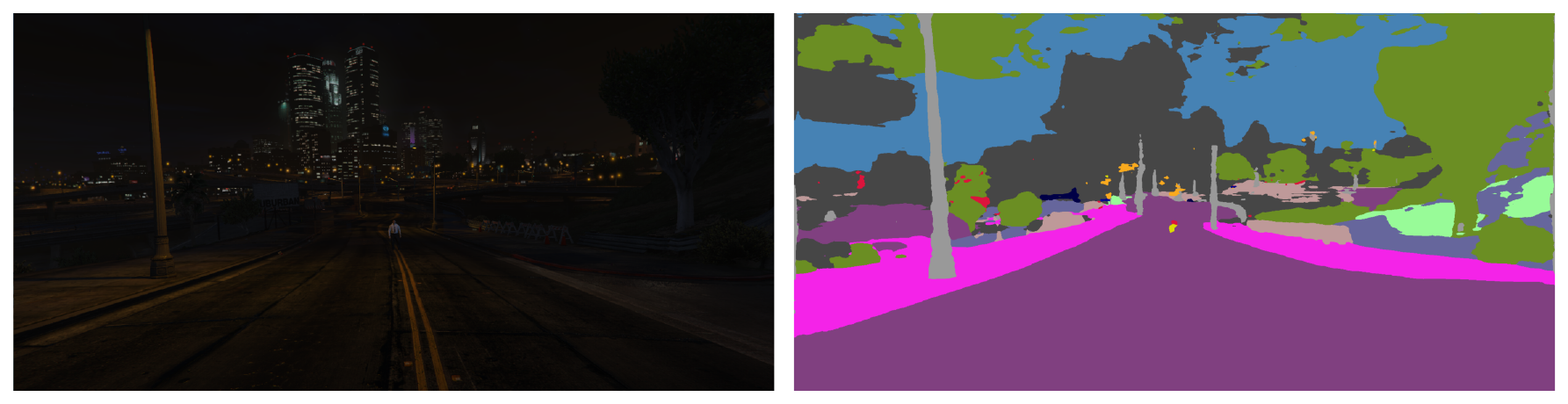}
    \end{subfigure}
    \begin{subfigure}{0.3\textwidth}
        \centering
    \includegraphics[width=\textwidth]{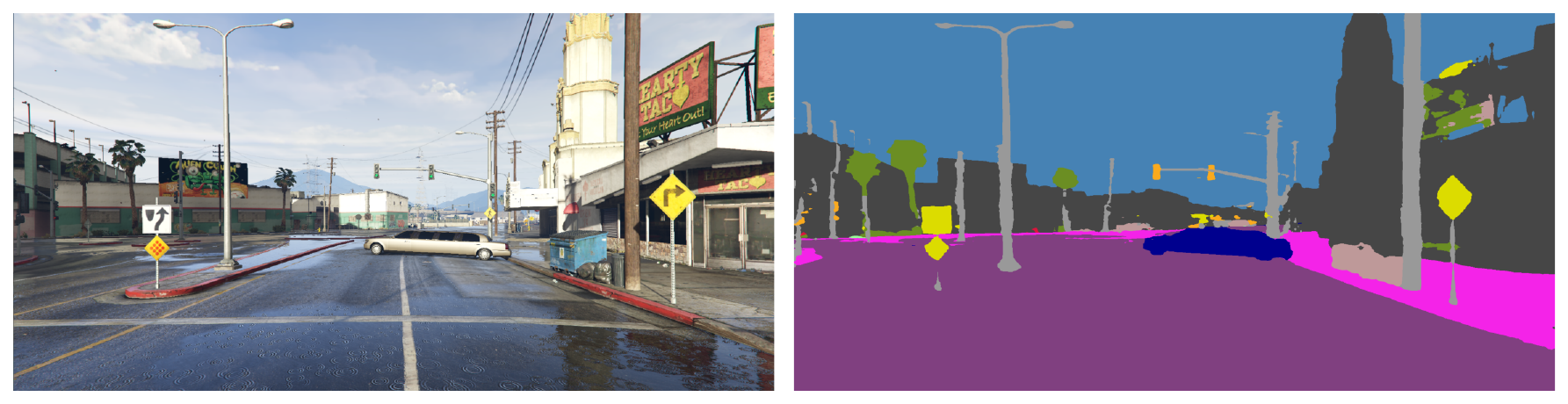}
    \end{subfigure}
    \begin{subfigure}{0.3\textwidth}
        \centering
    \includegraphics[width=\textwidth]{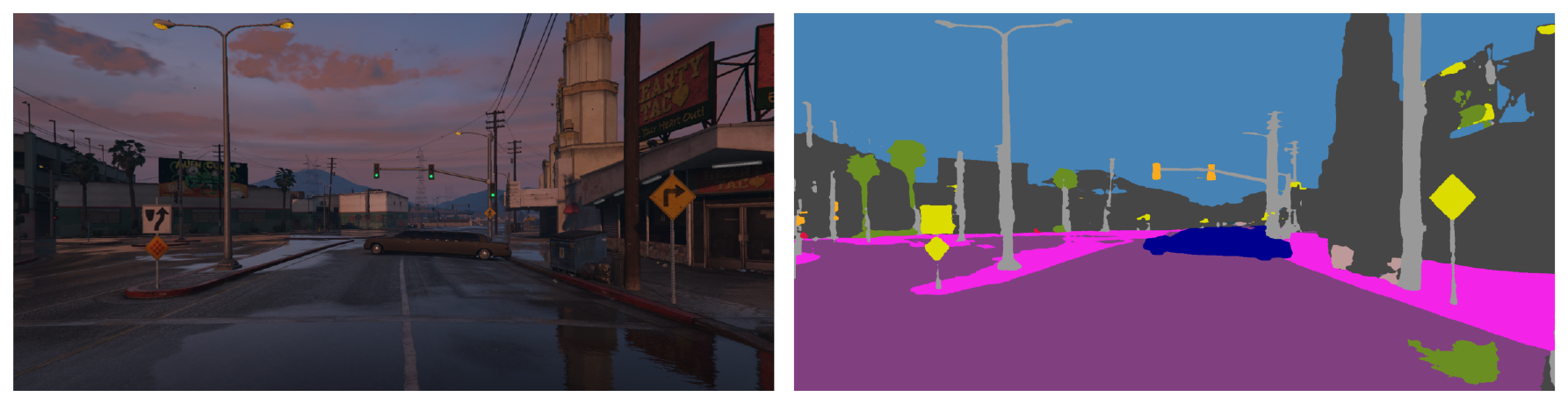}
    \end{subfigure}
    \begin{subfigure}{0.3\textwidth}
        \centering
    \includegraphics[width=\textwidth]{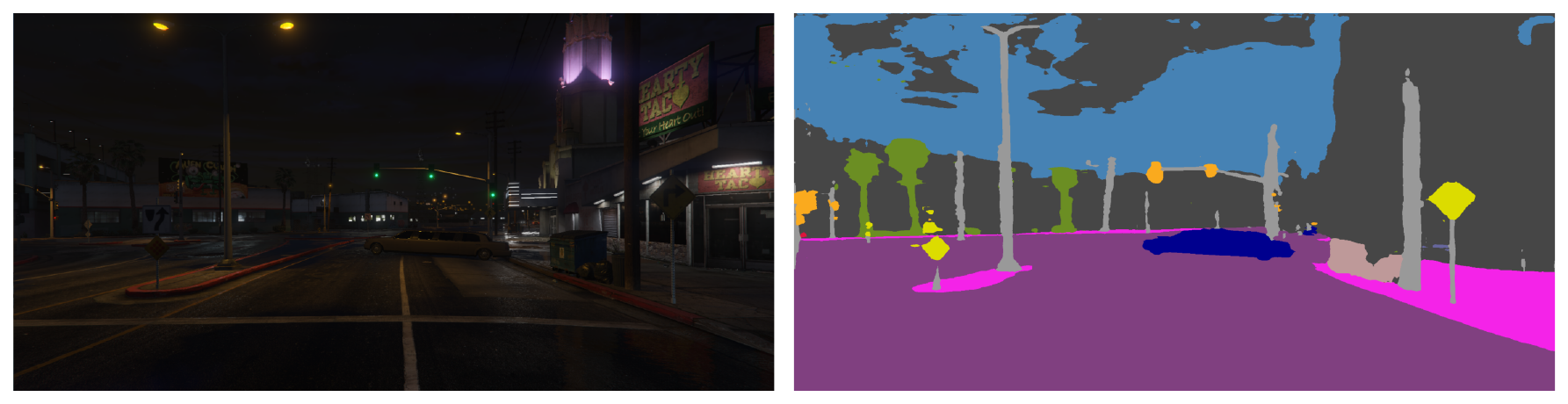}
    \end{subfigure}
    \begin{subfigure}{0.3\textwidth}
        \centering
    \includegraphics[width=\textwidth]{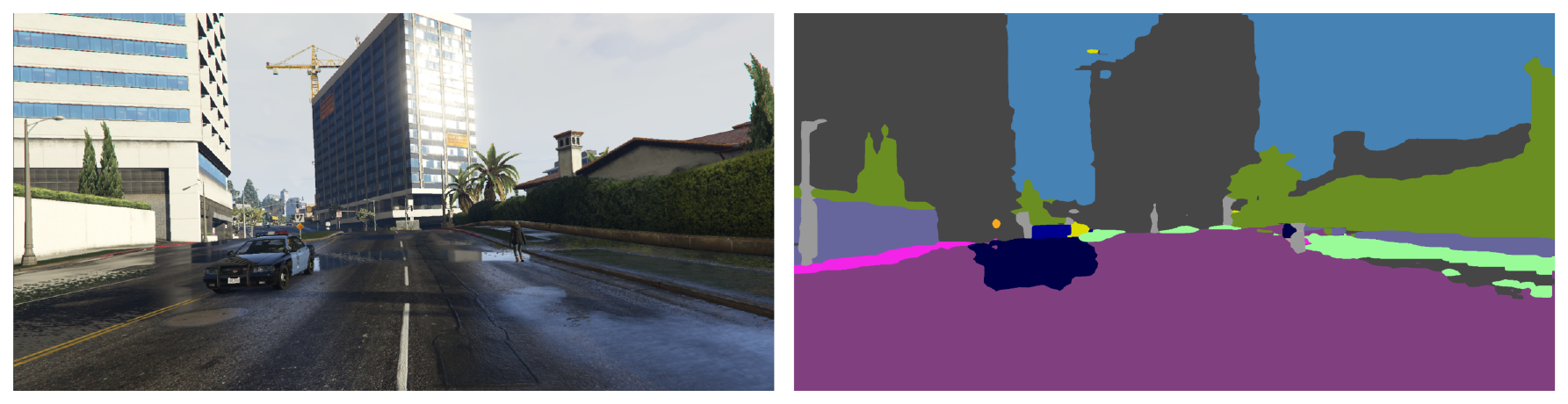}
    \end{subfigure}
    \begin{subfigure}{0.3\textwidth}
        \vspace{-0.5em}
        \centering
        \caption*{\footnotesize Mask2Former-Swin-L}
    \includegraphics[width=\textwidth]{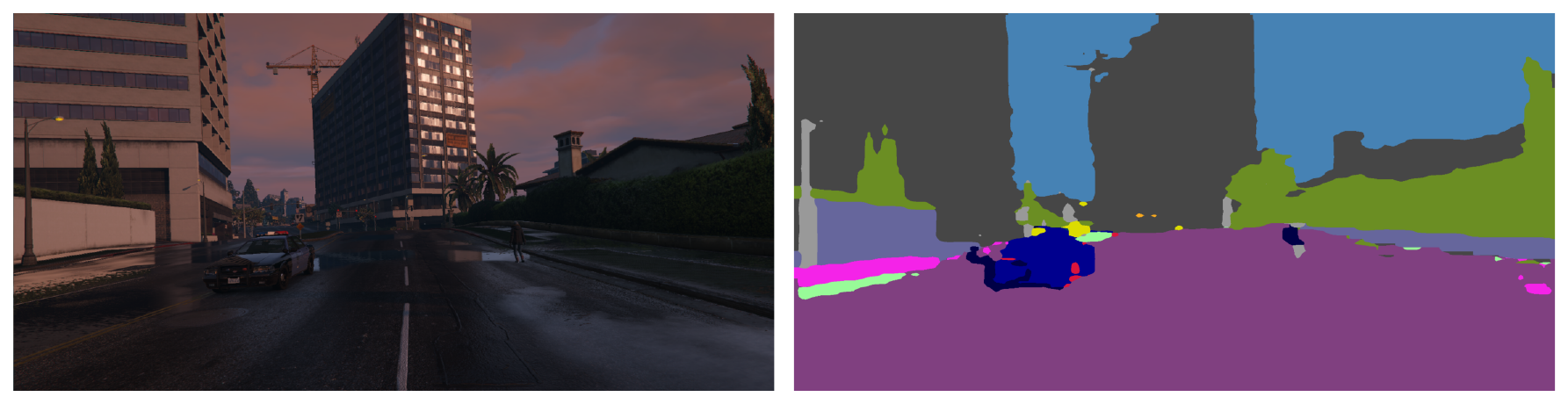}
    \end{subfigure}
    \begin{subfigure}{0.3\textwidth}
        \centering
    \includegraphics[width=\textwidth]{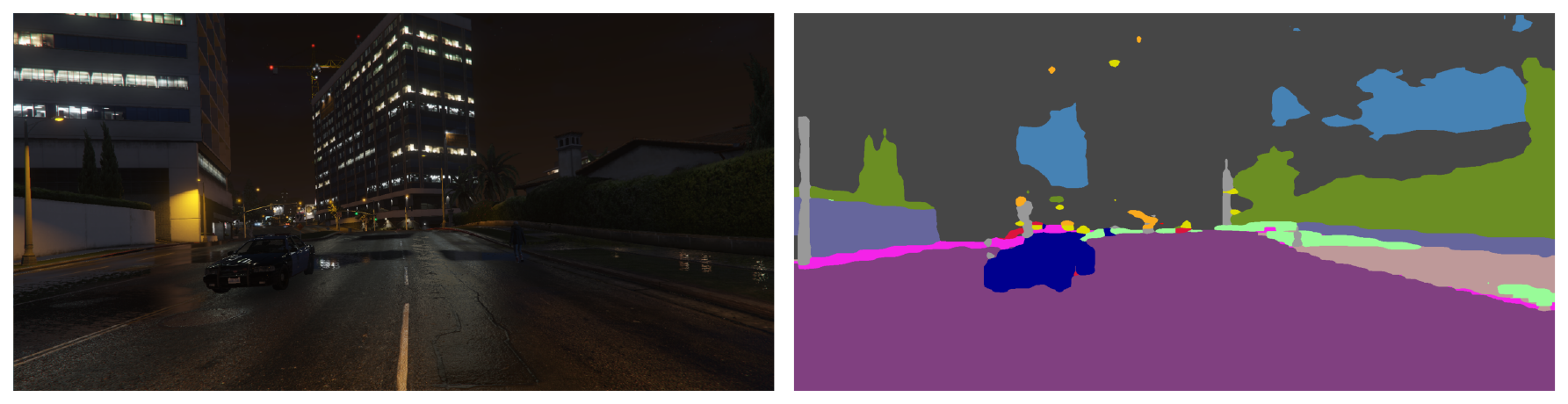}
    \end{subfigure}
    \begin{subfigure}{0.3\textwidth}
        \centering
    \includegraphics[width=\textwidth]{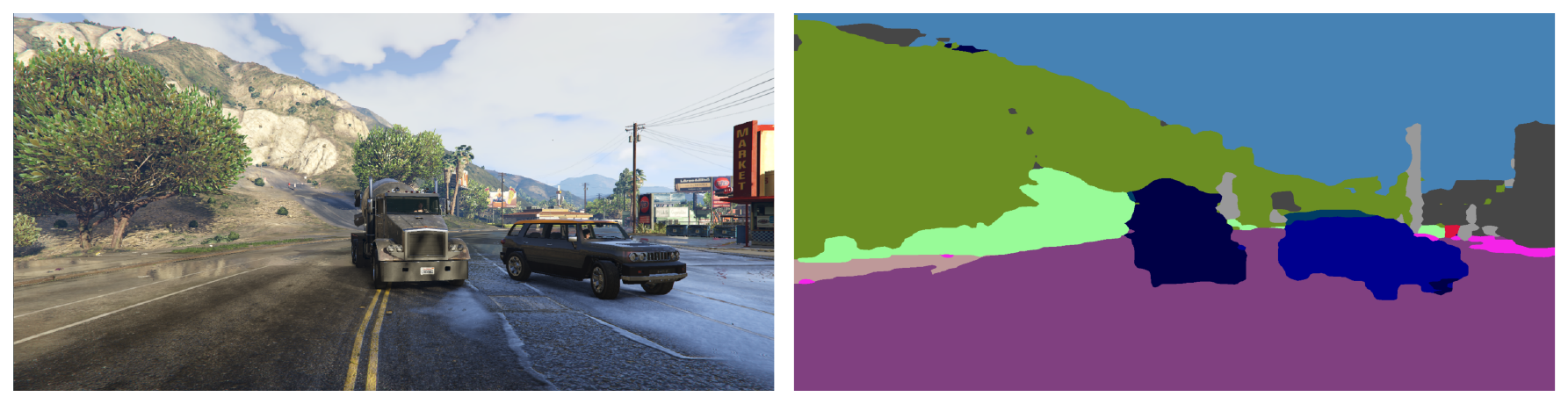}
    \end{subfigure}
    \begin{subfigure}{0.3\textwidth}
        \centering
    \includegraphics[width=\textwidth]{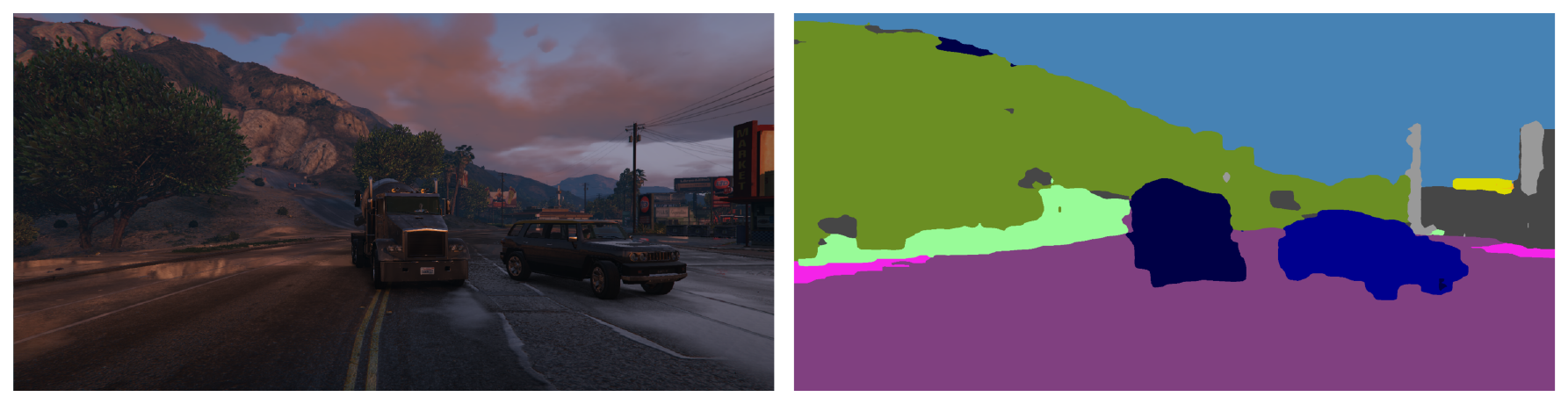}
    \end{subfigure}
    \begin{subfigure}{0.3\textwidth}
        \centering
    \includegraphics[width=\textwidth]{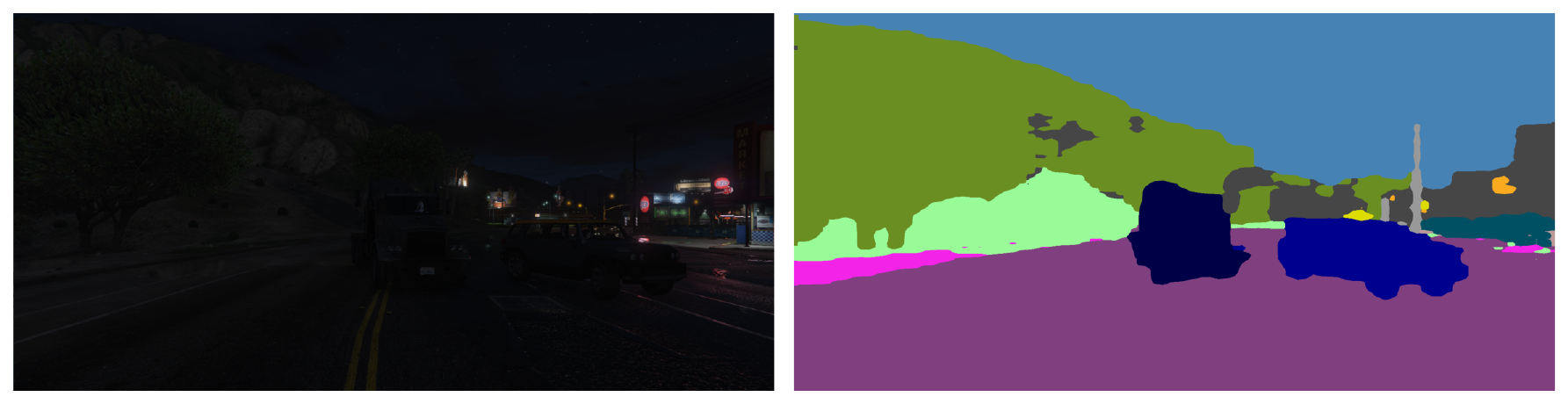}
    \end{subfigure}
        \caption{\small{Illustration of photometric shifts across three sunny scenarios under different illumination conditions: day, sunset, and night.}}
        \label{fig:qual_segformer}
\end{figure}

Table~\ref{tab:retention_conditions} reports class-wise reference consistency scores under three environmental shifts, using pseudo-labels extracted from SegFormer-B5 on the \textit{Day--Sunny} reference condition, as discussed above. 

For the \textit{Day--Sunny} $\rightarrow$ \textit{Day--Rain} shift, most models preserve stable predictions for large scene classes such as road, building, vegetation, sky, and car. SegFormer-B5 achieves the highest mean score, although this result may be partially biased by the use of pseudo-labels extracted from the same model. It is followed closely by SegFormer-B0, SegFormer-B2, and Mask2Former Swin-L. However, smaller or less frequent classes, such as traffic light, truck, and bicycle, show larger variability across models.

Under the stronger illumination shift \textit{Day--Sunny} $\rightarrow$ \textit{Sunset--Sunny}, the models still follow a similar trend. This is expected, as the GTA sunset scenario remains relatively close to the Cityscapes training distribution of the evaluated models. However, convolutional and lightweight models are more affected. While dominant classes remain relatively stable, categories such as terrain, rider, motorcycle, and bicycle are less consistently preserved, indicating that illumination changes affect fine-scale and dynamic objects more severely.

As expected, the most challenging setting is \textit{Day--Sunny} $\rightarrow$ \textit{Night--Sunny}, where all models exhibit a clear drop in mean consistency. SegFormer-B5 performs best on average, followed by Mask2Former Swin-L, Mask2Former Swin-S, and SegFormer-B2. The qualitative examples in Figure~\ref{fig:qual_segformer} further support these observations. Across different scenarios, SegFormer and Mask2Former preserve segmentation for large static regions, such as road, building, vegetation, and sky, under moderate changes from day to sunset. However, the night condition introduces substantial losses in visibility and local contrast, leading to less stable segmentation around small objects, boundaries, and distant scene elements.

\subsection{Analysis of photometric disentanglement}
\label{sec:ranking_analysis}

\begin{figure}[t]
\centering
    \begin{subfigure}{0.98\textwidth}
    \centering
    \begin{subfigure}{0.95\textwidth}
        \caption{\footnotesize{Our dataset.}}
        \label{fig:ranking_our_dataset}
        \begin{subfigure}{0.5\textwidth}
            \centering
            \includegraphics[width=\textwidth,trim=0mm 7.mm 0mm 0mm, clip=true]{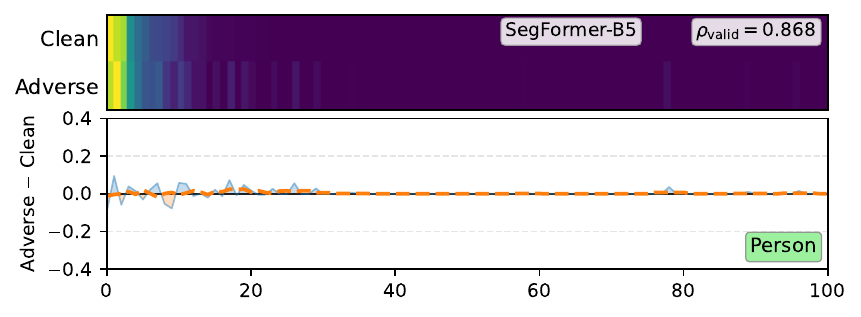}
        \end{subfigure}
        \begin{subfigure}{0.5\textwidth}
            \centering
            \includegraphics[width=\textwidth,trim=0mm 7.mm 0mm 0mm, clip=true]{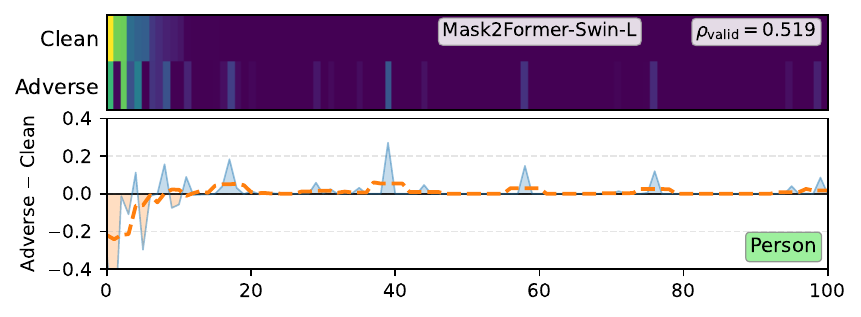}
        \end{subfigure}
        \begin{subfigure}{0.5\textwidth}
            \centering
            \includegraphics[width=\textwidth,trim=0mm 7.mm 0mm 0mm, clip=true]{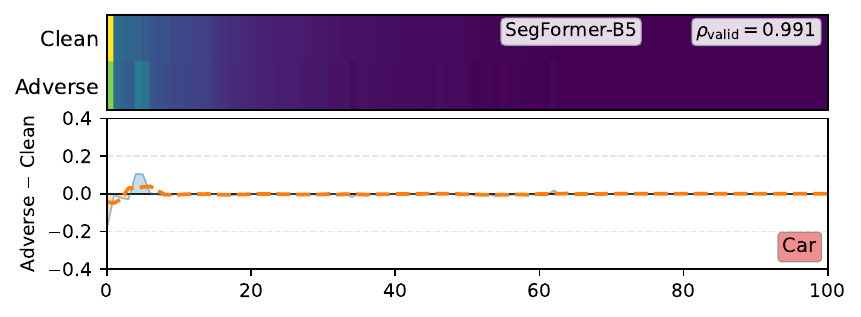}
        \end{subfigure}
        \begin{subfigure}{0.5\textwidth}
            \centering
            \includegraphics[width=\textwidth,trim=0mm 7.mm 0mm 0mm, clip=true]{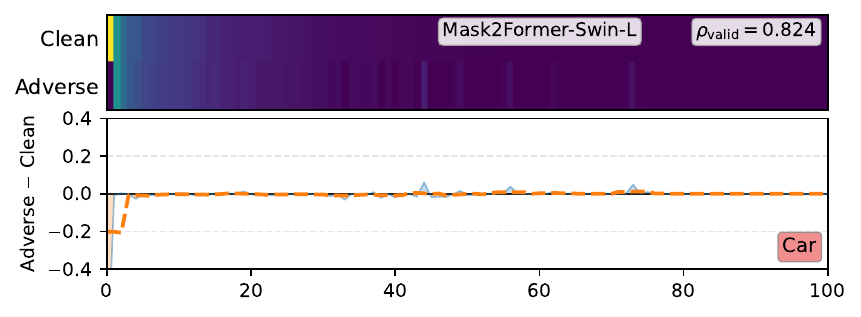}
        \end{subfigure}
    \end{subfigure}
    \begin{subfigure}{0.95\textwidth}
        \caption{\footnotesize{ACDC dataset.}}
        \label{fig:ranking_acdc_dataset}
        \begin{subfigure}{0.5\textwidth}
            \centering
            \includegraphics[width=\textwidth,trim=0mm 7.mm 0mm 0mm, clip=true]{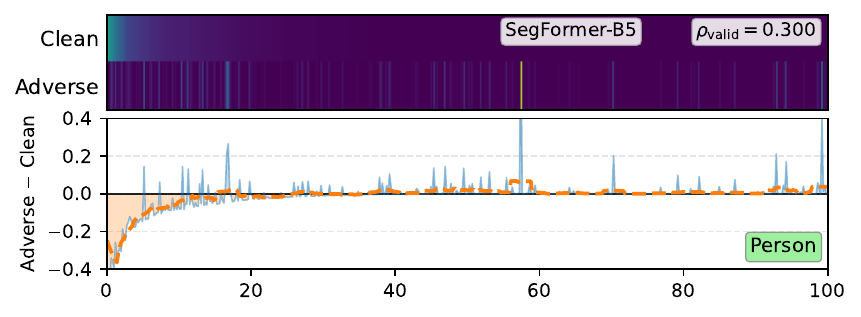}
        \end{subfigure}
        \begin{subfigure}{0.5\textwidth}
            \centering
            \includegraphics[width=\textwidth,trim=0mm 7.mm 0mm 0mm, clip=true]{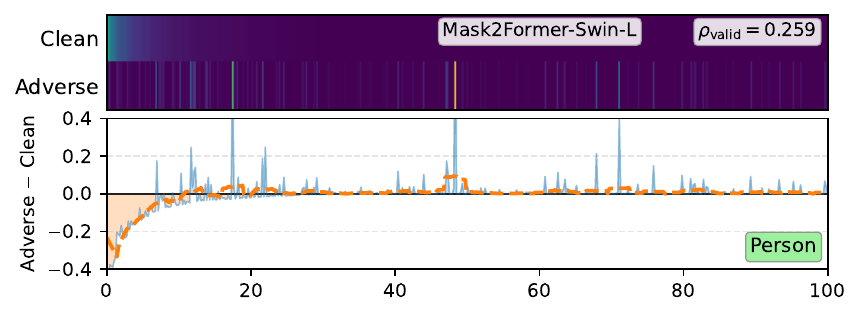}
        \end{subfigure}
        \begin{subfigure}{0.5\textwidth}
            \centering
            \includegraphics[width=\textwidth,trim=0mm 0mm 0mm 0mm, clip=true]{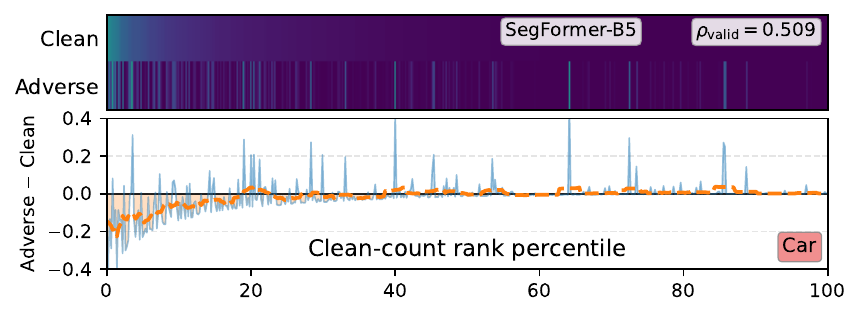}
        \end{subfigure}
        \begin{subfigure}{0.5\textwidth}
            \centering
            \includegraphics[width=\textwidth,trim=0mm 0mm 0mm 0mm, clip=true]{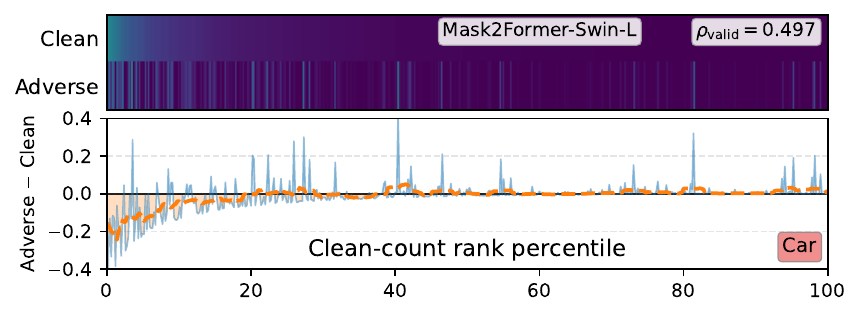}
        \end{subfigure}
    \end{subfigure}
    \end{subfigure}
    \caption{\small{Consistency analysis on our dataset and ACDC~\cite{sakaridis2021acdc}, using SegFormer (left) and MaskFormer (right). Frames are ranked according to the number of high-confidence predictions in clean scenarios for each class. Higher $\rho_{\mathrm{valid}}$ indicates stronger preservation of the clean-to-adverse ranking.}}
    \label{fig:consistency_comparisons}
\end{figure}

\begin{figure}[t]
\centering
\begin{subfigure}{0.95\textwidth}
    \centering
    \begin{subfigure}{0.48\textwidth}
        \centering
        \includegraphics[width=0.95\textwidth]{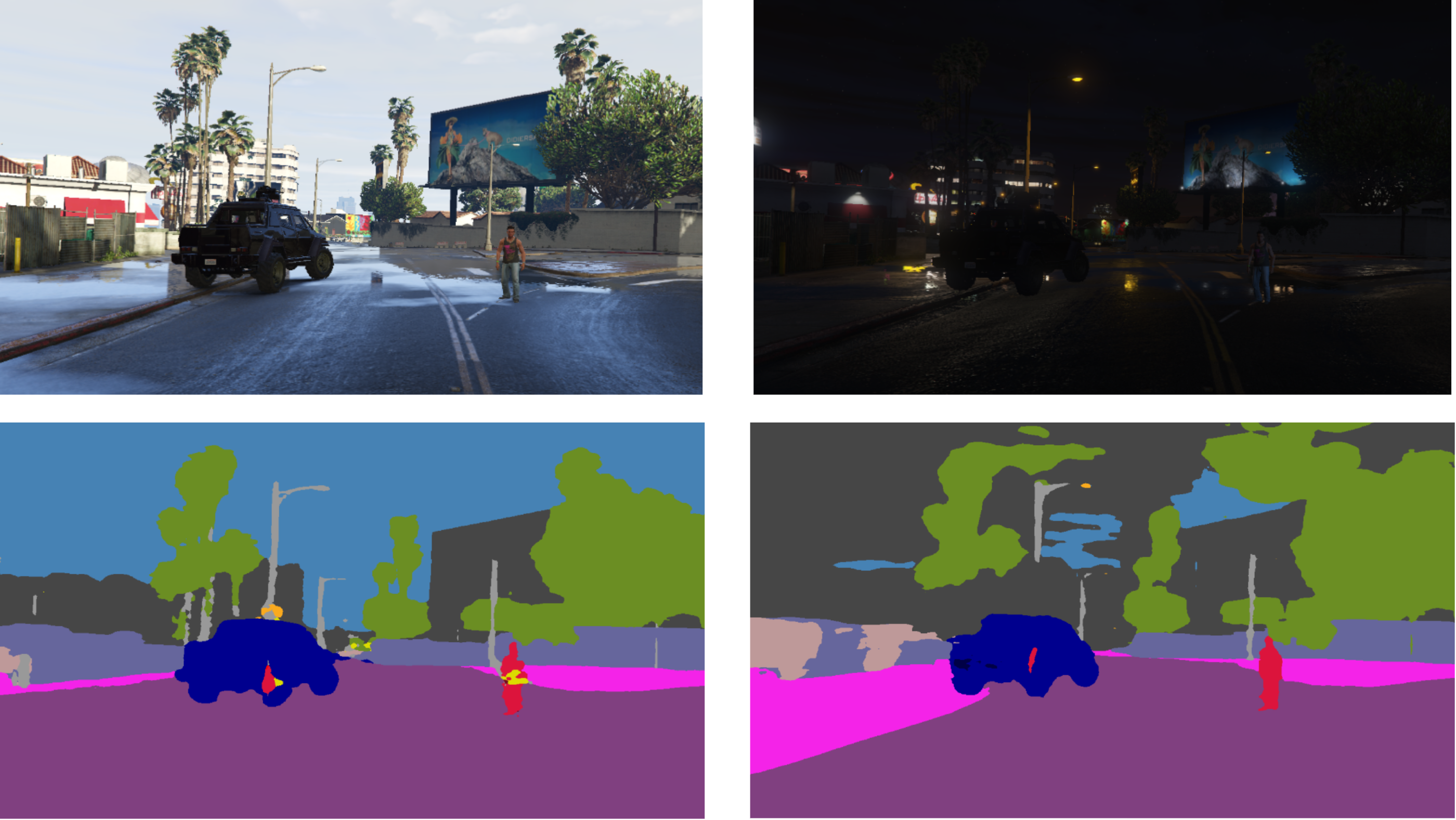}
    \end{subfigure}
    \begin{subfigure}{0.48\textwidth}
        \centering
        \includegraphics[width=0.95\textwidth]{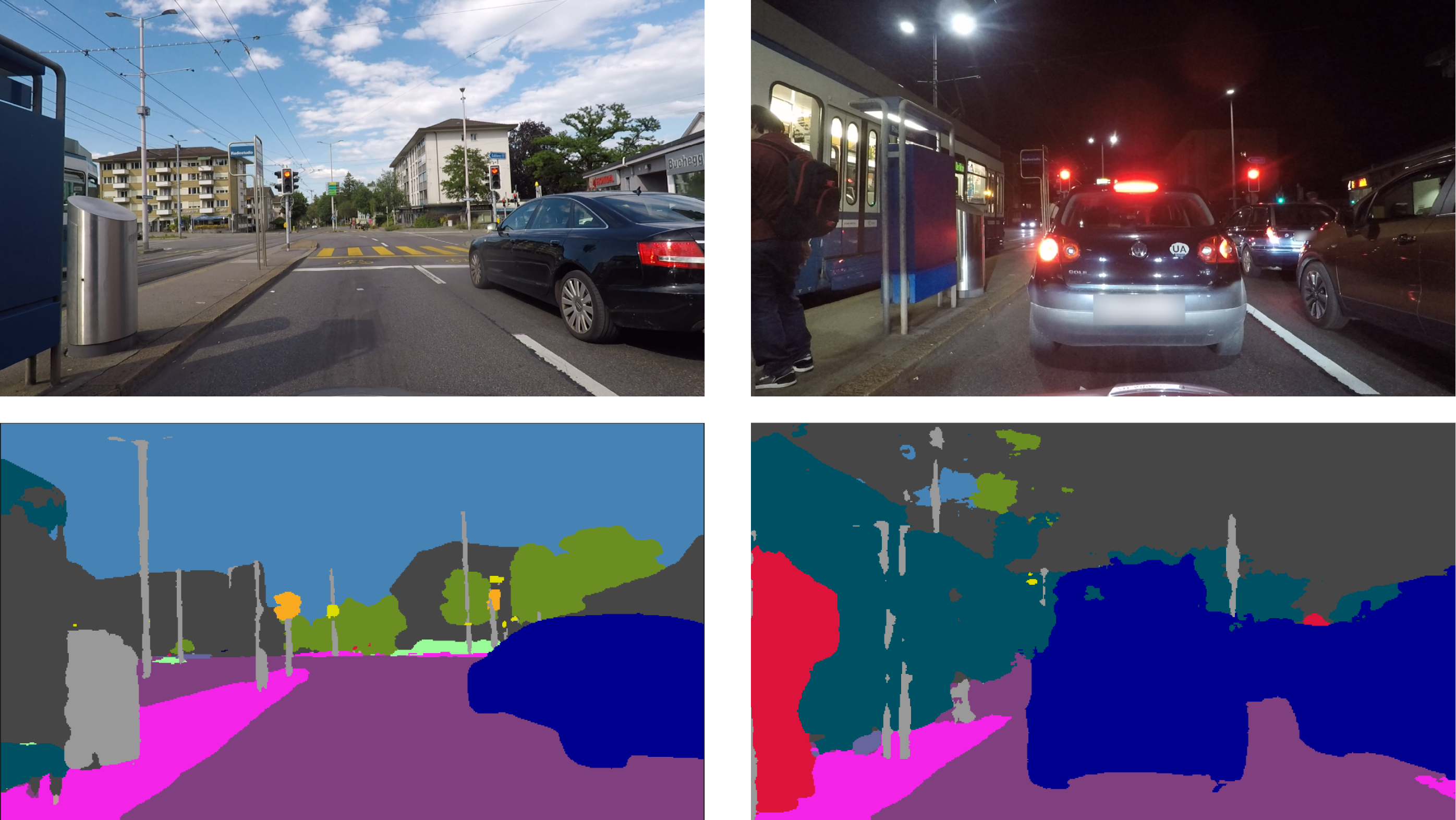}
    \end{subfigure}
    \end{subfigure}
    \caption{\small{Low-consistency examples for the \textit{person} class from our dataset, on the left, and ACDC, on the right, with segmentation outputs produced by SegFormer-B5.}}
    \label{fig:diff_acdc_our}
\end{figure}
To further highlight the importance of isolating photometric shifts during evaluation, we compare our dataset with ACDC~\cite{sakaridis2021acdc}, a coarsely paired real-world adverse-condition dataset commonly used for unsupervised domain adaptation and curriculum learning. In our dataset, each adverse sample is generated from the same reference scene, preserving camera pose, geometry, and object layout. As a result, changes in the model predictions can be more directly attributed to illumination and weather variations. Conversely, real-world paired datasets may also contain changes in dynamic objects, scene composition, and spatial arrangement, which can confound the analysis of photometric robustness.

To quantify this effect across the entire dataset, Figure~\ref{fig:consistency_comparisons} analyzes the stability of the high-confidence prediction mass for two dynamic classes, \textit{person} and \textit{car}. We consider a sunny image as the clean reference condition and a night-sunny image as the adverse condition, both for ACDC and for our dataset. For a generic image $\mathbf{X}_{k,i}$ and class $c$, we count the number of pixels predicted as class $c$ with high confidence, using the threshold $\tau=0.9$:
\begin{equation}
    \phi^{c}(\mathbf{X}_{k,i}) =
    \sum_{u}
    \mathbf{1}\!\left[
    \hat{y}_{k,i}^{(u)}=c
    \land
    P_{k,i}^{(u)}(c) \geq \tau
    \right],
    \label{eq:prediction_mass}
\end{equation}
where $\hat{y}_{k,i}^{(u)}$ is the predicted label at pixel location $u$ for image $\mathbf{X}_{k,i}$, and $P_{k,i}^{(u)}(c)$ is the predicted probability assigned to class $c$ at the same pixel.
Ideally, for a model invariant to photometric shifts and evaluated on perfectly paired images, the high-confidence prediction mass should be largely preserved between the clean and adverse conditions, leading to limited differences between $\phi^{c}(\mathbf{X}_{k,1})$ and $\phi^{c}(\mathbf{X}_{k,i})$. In practice, however, a variation in this quantity may indicate either poor model generalization to the photometric shift or imperfect correspondence between the paired images. In a fully paired dataset, the latter factor is minimized, making prediction differences more directly interpretable as model sensitivity to the adverse condition. In contrast, in coarsely paired real-world datasets, changes in object presence, object scale, or camera alignment may also alter the prediction mass independently of the photometric degradation.

In Figure~\ref{fig:consistency_comparisons}, frames $k$ are sorted according to the reference count $\phi^{c}(\mathbf{X}_{k,1})$ for visualization purposes. The same frame ordering is then used to inspect the corresponding adverse counts in the lower part of the heatmap. Since many frames may contain no, or only very few, high-confidence predictions for a given class $c$, computing the correlation over all  frames could artificially increase the consistency score due to uninformative zero-count samples. Therefore, we compute the ranking consistency only on the subset of frames that are most relevant for class $c$.
In particular, let $\mathcal{D}^{c}_{\mathrm{valid}} \subset \mathcal{D}$ denote the set of the top-$M$ frames with the largest reference counts $\phi^{c}(\mathbf{X}_{k,1})$, with $M=25$. We then define the valid-frame Spearman correlation as
\begin{equation}
    \rho_{\mathrm{valid}}^{c} =
    \mathrm{corr}_{\mathrm{Spearman}}
    \left(
    \left\{\phi^{c}(\mathbf{X}_{k,1})\right\}_{\mathbf{X}_{k} \in \mathcal{D}^{c}_{\mathrm{valid}}},
    \left\{\phi^{c}(\mathbf{X}_{k,a})\right\}_{\mathbf{X}_{k} \in \mathcal{D}^{c}_{\mathrm{valid}}}
    \right),
    \label{eq:valid_spearman}
\end{equation}
where $\mathbf{X}_{k,1}$ is the clean reference image for location $k$, and $\mathbf{X}_{k,a}$ is the corresponding adverse version. This score measures whether the frames that contain many confident predictions for class $c$ in the clean condition preserve a similar relative ordering under the adverse condition. In addition, the plots below the heatmaps report the image-wise clean--adverse difference (blue line) and its average trend (orange line).

The results in Figure~\ref{fig:consistency_comparisons} show that, in our dataset, the clean and adverse distributions remain largely aligned, particularly for the \textit{car} class. For this class, SegFormer-B5 
reaches $\rho_{\mathrm{valid}}^{c}=0.991$, while Mask2Former Swin-L reaches 
$\rho_{\mathrm{valid}}^{c}=0.824$. This indicates that frames containing many confident predictions in the reference condition tend to remain informative after the photometric shift. The \textit{person} class is less stable, as expected for a smaller and more visibility-sensitive category. Nevertheless, the correspondence between clean and adverse profiles remains substantially clearer than in ACDC.
Indeed, in ACDC, the same analysis leads to markedly lower correlations. For instance, for the \textit{person} class, the correlation drops to $\rho_{\mathrm{valid}}^{c}=0.300$ for SegFormer-B5 and $\rho_{\mathrm{valid}}^{c}=0.259$ for Mask2Former Swin-L, while the \textit{car} class also shows reduced consistency. These larger clean--adverse deviations suggest that the paired samples are affected not only by photometric degradation, but also by changes in object presence, scale, and spatial layout. This supports the observation that uncontrolled real-world pairing can make it difficult to disentangle true photometric robustness from broader scene-level variability. By contrast, our dataset provides a cleaner setting for evaluating robustness to illumination and weather shifts. Since geometry and scene content are preserved, prediction changes can be interpreted primarily as the effect of photometric perturbations, rather than as a consequence of uncontrolled semantic or spatial differences.

The analysis also reveals different behaviors across the two segmentation models. SegFormer generally preserves a more consistent clean--adverse ranking, whereas Mask2Former exhibits larger deviations under adverse conditions. This suggests that the proposed evaluation protocol can expose model-specific robustness profiles, complementing the aggregate class-level consistency analysis reported in Table~\ref{tab:retention_conditions}.
This interpretation is further supported by Figure~\ref{fig:diff_acdc_our}, where we inspect image pairs associated with low clean--adverse consistency for the \textit{person} class. In ACDC, the predictions may differ substantially not only because of the adverse condition, but also because the paired images contain different dynamic objects and are not perfectly aligned. Even small differences in camera position can prevent reliable pixel-wise correspondence between the clean and adverse samples. In contrast, in our GTA-based pairs, the scene geometry and object layout are preserved, so discrepancies in the output can be more directly associated with model behavior under the photometric shift. 

\begin{figure}[t]
        \vspace{1em}
        \centering
        \begin{subfigure}{0.75\columnwidth}
        \includegraphics[width=\linewidth]{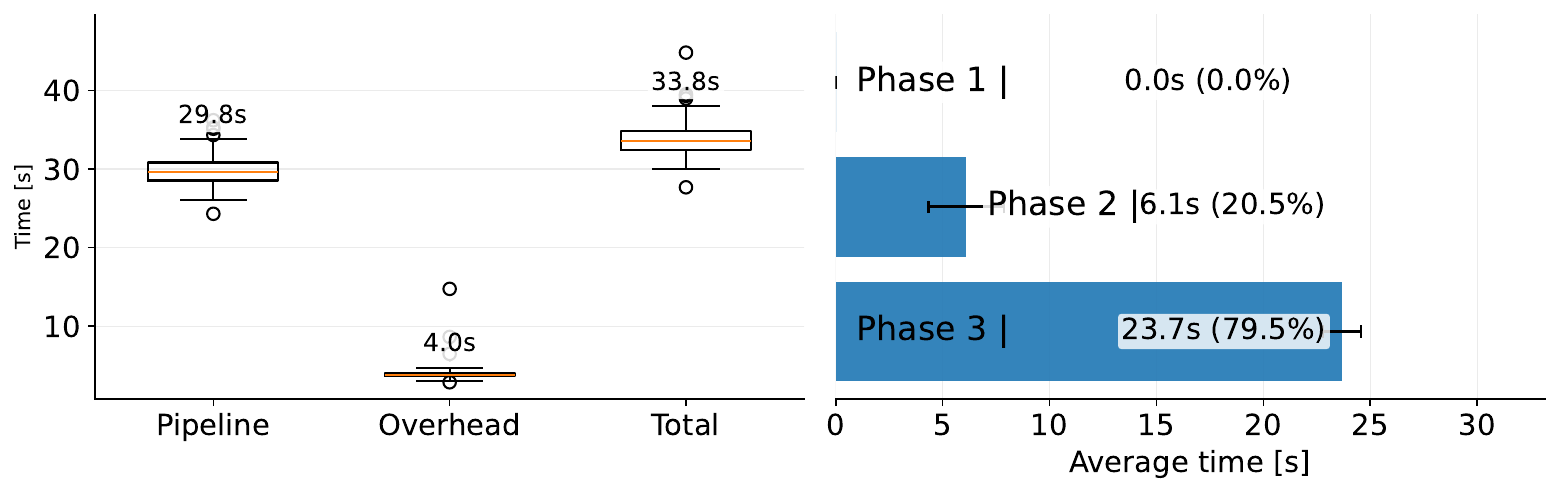}
        \end{subfigure}
        \caption{\small{Computational cost analysis of the proposed framework. Left: execution time over 100 scenarios for the pipeline, external overhead, and total process. Right: average pipeline-time decomposition across phases.}}
        \label{fig:time_analysis}
\end{figure}

\subsection{Cost analysis of the generation steps}

This section reports a computational cost analysis for generating paired images with the proposed framework. The experiments were conducted on a workstation equipped with an NVIDIA GeForce RTX 2080 GPU, 32 GB of RAM, and Windows 10, to evaluate the computational cost without relying on expensive GPUs or dedicated servers. In particular, the generation procedure described in Algorithm~\ref{alg:dataset_generation} was tested across 100 scenarios. For each scenario, we produced nine versions, considering the weather and environmental conditions discussed in Section~\ref{sec:framework}.

Figure~\ref{fig:time_analysis} reports the execution time analysis. The left plot shows the distribution of the computational time across the 100 scenarios, distinguishing between the time required by the proposed pipeline, the external overhead introduced by connection and debugging functions, and the resulting total time. On average, the pipeline requires 29.8 s per scenario, while the external overhead accounts for only 4.0 s. The average total execution time is therefore 33.8 s. The boxplots also show that the execution time remains relatively stable across scenarios, with limited variability for both the pipeline and the total time. A few outliers are present, mainly in the overhead and total time, but they do not substantially affect the overall trend.

The right plot further decomposes the average pipeline time into the different phases discussed in Section~\ref{sec:framework}. Phase 1 has a negligible computational impact, while Phase 2 requires 6.1 s, corresponding to 20.5\% of the pipeline time. As expected, Phase 3 is the most computationally demanding step, requiring 23.7 s and accounting for 79.5\% of the pipeline time. This is because Phase 3 includes the generation of all adverse conditions considered in our experiments. However, if only a limited subset of conditions is required, the computational cost can be significantly reduced.

\section{Conclusions}
\label{sec:conclusion}
In this work, we presented \NAME, a framework for generating paired images under controlled photometric shifts. By leveraging a simulated environment, the proposed pipeline renders the same underlying scene across different illumination and weather conditions while preserving camera pose, geometry, and scene composition. This enables the effect of photometric changes to be studied in isolation, reducing the confounding factors that typically affect real-world adverse-condition datasets.
We used the generated data to evaluate semantic segmentation models under different environmental conditions. The results show that photometric shifts alone can affect model predictions, with degradation patterns varying across models, conditions, and classes, while at the same time avoiding inconsistency drop due to the semantic variations.

Overall, the proposed framework provides a controlled testbed that complements real-world benchmarks. 
We believe that the proposed technique for constructing synthetic datasets can contribute to the development and improvement of autonomous driving systems, particularly when combined with real-world data and sim-to-real approaches.

\subsection{Limitations and Future Directions}
The proposed framework enables the generation of paired images under controlled photometric shifts, but it also presents some limitations. First, the interaction with the simulation environment is inherently indirect. We do not have access to the source code of the GTA V game engine or to the internal values of its native function calls. Instead, the framework relies on third-party tools, such as DeepGTAV, VPilot, and Script Hook V, which expose a game-level interface to control and query the environment. While this abstraction makes the pipeline feasible, it limits the degree of control over scene elements, environmental parameters, and the range of modifications that can be applied.

As a result, some information useful for dataset generation, such as exact spatial measurements, object dimensions, and distances, is only partially accessible and may require indirect estimation. Although our pipeline mitigates these issues through heuristic strategies, including approximate geometric reasoning and distance estimation, game-based environments can still introduce variability. For example, objects may occasionally appear in unexpected positions, stochastic events may occur within the game engine or the communication server, and additional inconsistencies may arise from bugs or unintended behaviors in the external tools. For this reason, a manual verification step is still suggested to identify and discard potential anomalous samples.

A further limitation is the lack of direct access to perfect semantic ground truth, as discussed in Section \ref{sec:framework}. Unlike simulation platforms specifically designed for autonomous-driving research, the game environment used in this work does not provide dense semantic labels. Therefore, our analysis focuses on output consistency and prediction changes with respect to clean sunny reference images, which could be considered as weak ground truth, rather than on absolute performance against simulator-provided annotations. This choice is also aligned with recent game-based works that rely on pseudo-labels or model-based references when ground truth is not directly available. In our case, the clean reference condition provides a meaningful comparison point to study how model outputs change when only photometric conditions are modified.

Future work will focus on improving the automatic validation of generated samples, extending the framework to additional perception tasks, and studying the sim-to-real relevance of the observed robustness trends. We also plan to investigate the integration of generative AI models with game-based simulation, using them to support anomaly detection, scene refinement, and the generation of richer environmental variations.

{
\small
\section*{Acknowledgements}
This work was partially supported by project SERICS (PE00000014) under the MUR (Ministero dell'Università e della Ricerca) National Recovery and Resilience Plan funded by the European Union - NextGenerationEU.
}

\bibliographystyle{unsrt}
\bibliography{main}
\end{document}